\documentclass{article}
\usepackage{spconf,amsmath,graphicx}
\usepackage{multirow}
\usepackage{amsfonts}

\usepackage{enumitem}
\setlist{nosep, leftmargin=14pt}

\usepackage{mwe} 

\title{Evaluating and Mitigating Skin Color Bias in Skin Lesion Classification}
\name{Kuniko Paxton, Koorosh Aslansefat, Dhavalkumar Thakker, Yiannis Papadopoulos \thanks{Kuniko Paxton, Koorosh Aslansefat, Dhavalkumar Thakker, Yiannis Papadopoulos}}
\address{School of Digital and Physical Sciences, University of Hull}
\begin{document}

%
\maketitle

\raggedbottom
\begin{abstract}
Skin lesion classifiers may overlook performance disparity because of subtle skin tone variations by treating skin tone as a categorical attribute. This study proposes a method that represents skin tone as a continuous attribute and assigns individual weights to each sample based on its nonlinear relationship with classification performance. Evaluations on two dermatological datasets showed that fairness metrics were sensitive to skin-color categorization. The proposed method also demonstrated more stable fairness improvements than conventional categorical reweighting while generally preserving classification performance.
\end{abstract}
\section{Introduction}
\label{sec:intro}
While adopting automated diagnostic support technologies using machine learning is progressing in dermatology, ensuring that their performance does not vary significantly based on patient skin characteristics has become a critical challenge. Several studies have reported skin tone differences degrade classification performance, making fairness in skin lesion classification models a key research topic. Existing dermatological image datasets exhibit biases in skin tone distribution, such as a prevalence of light skin tones and a scarcity of dark skin tones \cite{liu2023artificial,groh2021evaluating, guo2022bias}. Prior studies evaluated disparities in classification performance related to skin tone and attempted to mitigate these biases \cite{bevan2022detecting,pakzad2022circle,sarridis2023towards}. 

Prior approaches to addressing skin tone bias in skin lesion classification can broadly be categorized into methods at the data level, model level, and evaluation level. Data-level methods supplement small skin tone groups using techniques such as image augmentation \cite{corbin2023assessing} and generation \cite{pakzad2022circle,pakzad2024addressing}. Model-level methods suppress the performance dependence on specific skin tone groups, such as reweighting, regularisation \cite{gaddey2024patchalign,xu2023fairadabn}, adversarial learning \cite{du2022fairdisco}, unlearning \cite{bevan2022detecting}, and pruning \cite{wu2022fairprune}. Evaluation-level methods comprise counterfactual methods \cite{narayanaswamy2021using} and bias mitigation techniques in the aggregation phase in federated learning \cite{fan2021fairness}. Despite the divergent stages at which these methodologies intervene, they are contingent on predefined skin tone groups, such as the Fitzpatrick skin scale \cite{fitzpatrick1988validity} or sensitive-attribute labels. 

However, we must consider the nature of skin tone, which is an inherently continuous attribute. Since skin tone varies between individuals and exhibits unevenness and nuance within a single image, exhibiting local variations in hue, a single categorical label may fail to fully capture continuity and within-image variation. Moreover, evaluators may label skin tone differently, and categorization may introduce new uncertainties and bias in grouping. The implications of dividing the inherently continuous attribute of skin color into discrete categories—and the resulting potential for overlooking underlying biases—have not been sufficiently examined. 

Our previous work \cite{paxton2024measuring} established the use of continuous skin-color representations for fairness assessment towards adversarial attacks. Building on this foundation, the present study introduces class-conditional Bayesian binomial spline regression to capture nonlinear variations in diagnostic performance across the skin-color spectrum. The resulting performance estimates are used to assign individual training weights based on diagnostic class and skin-color distance. We investigate whether this approach provides a more stable fairness–accuracy trade-off across two dermatological datasets and four neural-network architectures.

Main contributions in this research is following: (1) We proposes a method to measure performance fairness with respect to skin tone be quantified without categorization in skin lesion classification. (2) We demonstrate the sensitivity of group-based fairness metrics to skin-color categorization, highlighting the limitations of discrete grouping for continuous attributes. (3) We develop a bias-mitigation algorithms to improve fairness discovered by measuring skin tone continuously while maintaining overall classification performance by introducing class-conditional Bayesian binomial spline regression to model nonlinear variations in diagnostic performance.

\section{Methodology}
\begin{figure*}[!t]
    \centering
    \includegraphics[width=0.98\linewidth]{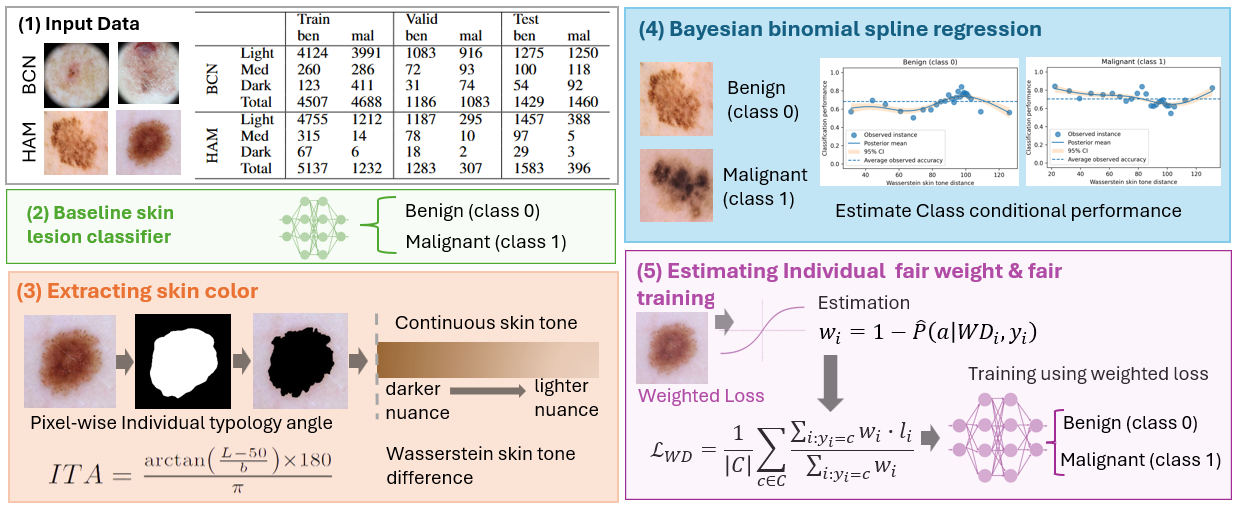}
    \caption{Proposed fairness evaluation and bias mitigation based on continuous skin tone attributes. Skin tone differences are represented using the Wasserstein distance, and each sample is individually weighted based on performance estimates derived from Bayesian binomial spline regression. Here, $a_i\in\{0,1\}$ denotes correctness and $\mathcal{C}$ the set of classes.}
    \label{fig:skin_overview}
\end{figure*}

Figure \ref{fig:skin_overview} illustrates the overall framework for measuring skin tone as a continuous attribute and mitigating the performance disparities identified through this representation.


\textbf{Measuring skin tone}: 
Skin tone was represented by the pixel-wise ITA distribution 
$D_i$ of each image. Its difference from a reference distribution 
$D_0$ was quantified by the 1-Wasserstein distance, $WD_i = W_1(D_0,D_i)$. \\

\textbf{Continuous fairness measurement}: 
Let $a_i=\mathbb{I}(\hat{Y}_i=Y_i)$ denote classification correctness,
and let $y_i\in\mathcal{C}=\{0,1\}$ denote the class label. 
We define the class-conditional correctness function as

\begin{equation}
    p_y(z)=P(a=1\mid Y=y,WD=z)
\end{equation}

The continuous performance disparity relative to the reference skin tone is
\begin{equation}
\Delta_y(z)
=
\frac{|p_y(z)-p_y(0)|}{z},
\qquad z>0.
\end{equation}

For $y=1$, $p_1(z)$ corresponds to TPR (Equal Opportunity), whereas for $y=0$,
$p_0(z)=1-\mathrm{FPR}(z)$; therefore, the absolute variation in
$p_0$ is equivalent to the corresponding FPR variation. \\

\textbf{Class-conditional performance estimation}: 
For each class $y$, $p_y(z)$ was estimated using Bayesian binomial
spline regression. Let $k_j$ and $n_j$ denote the number of correct
predictions and total samples in WD bin $j$, respectively:
\begin{equation}
k_j \sim
\mathrm{Binomial}
\left(
n_j,\,
\sigma(B(WD_j)^\top\beta_y)
\right).
\end{equation}

A Gaussian prior with a second-order smoothness penalty was imposed on
$\beta_y$, and the posterior was approximated using the Laplace approximation.
The posterior mean correctness probability for sample $i$ was
\begin{equation}
\hat p_i =
E_{\beta_{y_i}\mid\mathcal{V}}
\left[
\sigma\!\left(B(WD_i)^\top\beta_{y_i}\right)
\right],
\qquad
w_i=1-\hat p_i,
\end{equation}

where $\mathcal{V}$ denotes the validation data and $w_i$ is the
individual training weight. \\

\textbf{Evaluation}: Although the fairness evaluation in this study does not require a predefined skin tone category, we adopted the widely used Reweight method as a comparison method to ensure a fair comparison with existing category-based approaches to validate our methods.

\section{Experimental Setup}
\textbf{Datasets}: Experimental datasets are HAM10000 \cite{tschandl2018ham10000} and BCN20000 \cite{hernandez2024bcn20000} for binary classification (divided into benign and malignant) as Table in Figure \ref{fig:skin_overview}. The data were split at the lesion level to prevent images from the same lesion from appearing in different subsets, and the test set was held out for final evaluation. \textbf{Preprocessing to measure skin tone}: The lesion regions were masked using segmentation, and hair was removed using an existing morphological preprocessing method that combines morphological opening, black-hat transformation, CLAHE, and thresholding. The hyperparameters for each process were tuned to suit the dataset by human visual inspection. \textbf{Models}: ResNet50 \cite{he2016deep} as a conventional CNN, MobileNetV2 \cite{sandler2018mobilenetv2} and EfficientNet-B0 \cite{Tan2019EfficientNetRM} as lightweight CNNs, and Swin Transformer \cite{liu2021swin} as a transformer-based model were selected to assess the architectural dependency. Each model was fine-tuned for 50 epochs using ImageNet-pretrained weights across five random seeds, and for each seed, the checkpoint achieving the highest F1 score on the validation set was selected as the baseline.


\section{Results and Discussion}
\subsection{Impacts of categorization of skin color}
Table \ref{tab:baseline} shows the classification performance of baseline models that do not account for fairness, as well as the performance gaps under three- and six-group skin-tone categorizations. Across both datasets and all architectures, the six-group evaluation demonstrated larger gaps for all metrics.

In BCN, the smallest increase in the gap was 2.0 percentage points (pp) in the F1 gap for ResNet50, and the largest was 13.3 pp in the FPR gap for SwinT. In the HAM, the increase in the F1 gap was particularly pronounced, with increases ranging from 33.2 to 42.1 pp observed across all models. This significant change may be attributed to the fact that some skin tone groups had a small number of samples or classes, causing a few misclassifications to have a large impact on the group-level F1 score. Still, the increase in disparities was not limited to the F1 gap; it was consistently observed in the TPR gap, FPR gap, and Equalized-Odds gap as well.

These results highlight the sensitivity of fairness metrics to categorization. As illustrated in Figire \ref{fig:estimator}, diagnostic performance may vary nonlinearly across the continuous skin-tone representation, making categorical evaluations sensitive to group boundaries. These observations motivate fairness assessment and mitigation based on continuous attributes rather than relying solely on predefined categories.

\begin{table*}[!t]
\centering
\resizebox{0.9\textwidth}{!}{%
\begin{tabular}{ll|ll|ll|ll|ll|l}
\hline
&      & Acc & Gap $3G\to 6G$ & F1    & Gap $3G\to 6G$ & TPR & Gap $3G\to 6G$ & FPR & Gap $3G\to 6G$ & EOdds $3G\to 6G$ \\ \hline
\multirow{4}{*}{\rotatebox[origin=c]{90}{BCN}} & Res    & 0.734    & 8.1→11.2 (+3.1)  & 0.734 & 6.1→8.1 (+2.0)    & 0.749                   & 26.0→35.3 (+9.3)  & 0.281                   & 19.0→29.1 (+10.1) & 26.4→37.8 (+11.4)  \\
                     & Mob & 0.680    & 5.1→11.5 (+6.4)  & 0.680 & 4.6→11.4 (+6.9)   & 0.648                   & 9.9→20.4 (+10.5)  & 0.287                   & 11.7→19.9 (+8.1)  & 13.1→24.0 (+10.8)  \\
                     & Swi       & 0.780    & 6.7→10.3 (+3.6)  & 0.780 & 5.8→10.0 (+4.2)   & 0.735                   & 21.6→29.2 (+7.7)  & 0.174                   & 12.0→25.2 (+13.3) & 21.6→29.3 (+7.8)   \\
                     & Eff & 0.703    & 4.1→8.6 (+4.6)   & 0.703 & 4.0→9.8 (+5.8)    & 0.702                   & 9.0→15.0 (+6.0)   & 0.296                   & 12.7→23.8 (+11.1) & 14.0→24.7 (+10.8)  \\
\multirow{4}{*}{\rotatebox[origin=c]{90}{HAM}} & Res    & 0.774    & 16.9→25.2 (+8.3) & 0.712 & 5.2→47.1 (+41.9)  & 0.776                   & 16.1→26.3 (+10.2) & 0.227                   & 18.6→25.9 (+7.3)  & 18.6→26.3 (+7.8)   \\
                     & Mob & 0.766    & 17.4→24.6 (+7.2) & 0.675 & 12.0→48.7 (+36.7) & 0.592                   & 23.6→28.4 (+4.9)  & 0.191                   & 16.0→20.2 (+4.2)  & 26.3→29.2 (+3.0)   \\
                     & Swi       & 0.865    & 10.0→16.8 (+6.8) & 0.791 & 9.2→51.3 (+42.1)  & 0.674                   & 27.9→37.9 (+10.0) & 0.087                   & 5.9→11.1 (+5.2)   & 27.9→37.9 (+10.0)  \\
                     & Eff & 0.795    & 15.1→24.3 (+9.1) & 0.706 & 12.9→46.1 (+33.2) & 0.612 & 33.2→36.0 (+2.8)  & 0.159                   & 12.6→19.5 (+6.9)  & 35.4→39.0 (+3.7)   \\ \hline
Av & —           & 0.762    & 10.4→16.6 (+6.1) & 0.723 & 7.5→29.1 (+21.6)  & 0.686 & 20.9→28.6 (+7.7)  & 0.213                   & 13.6→21.8 (+8.3)  & 22.9→31.0 (+8.1)  \\
\hline
\end{tabular}
}
\caption{Baseline performance and intergroup gaps under 3-group and 6-group skin-tone categorization. Parentheses indicate the change from 3G to 6G in percentage points (pp).}
\label{tab:baseline}
\end{table*}

\subsection{Bayesian Performance Estimator}

\begin{figure}[!t]
    \centering
    \includegraphics[width=0.95\linewidth, trim=0mm 8mm 0mm 8mm, clip]{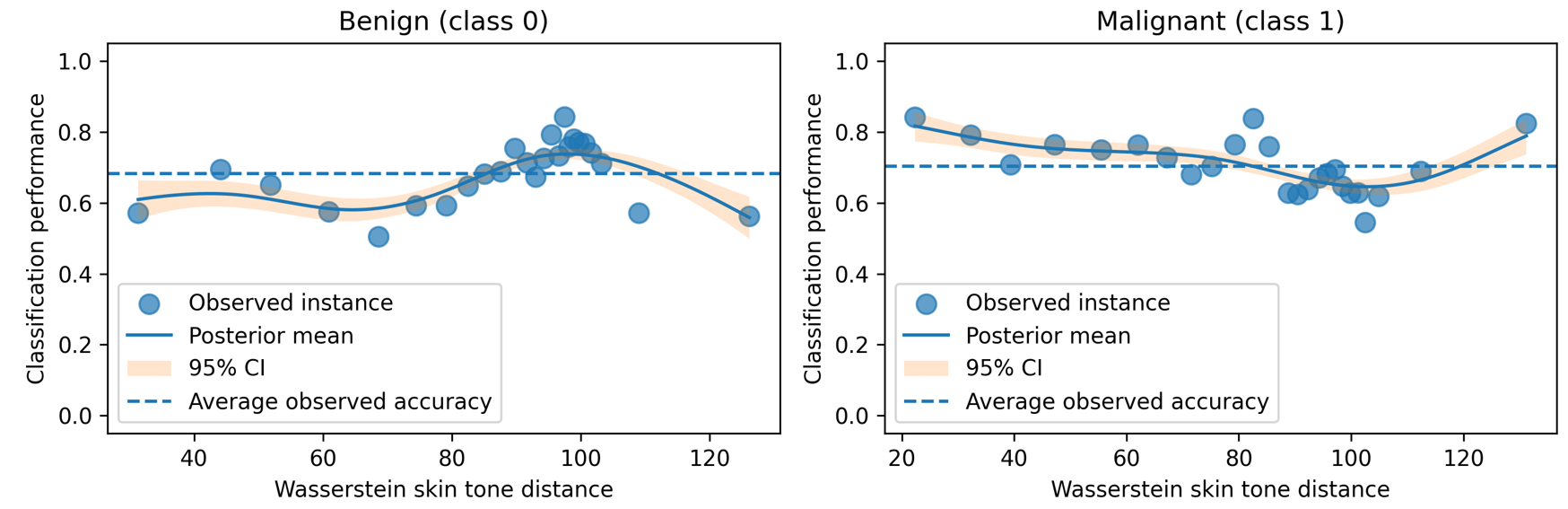}
    \caption{Sample results of estimating the relationship between skin tone attributes and classification performance for EfficientNet-B0 using Bayesian regression.}
    \label{fig:estimator}
\end{figure}

Figure \ref{fig:estimator} shows the results of estimating the relationship between continuous skin tone attributes and classification performance using Bayesian spline regression for each class. For both the “Benign” and “Malignant” classes, the classification performance did not vary monotonically or linearly with skin tone attributes but exhibited localized, class-dependent nonlinear patterns. The solid lines represent the posterior mean, while the shaded regions indicate pointwise 95\% credible intervals based on the Laplace approximation. These results suggest that when continuous skin tone attributes are divided into a small number of discrete categories, the local performance variations within those categories are averaged out, potentially leading to an underestimation of performance disparities related to skin tone. This is consistent with the difference in disparities observed between the three-group and six-group evaluations in Table \ref{tab:baseline}.

\subsection{Validation of Proposed Method}
\begin{figure*}[!t]
    \centering
    \includegraphics[width=0.98\linewidth, trim=0mm 8mm 0mm 8mm, clip]{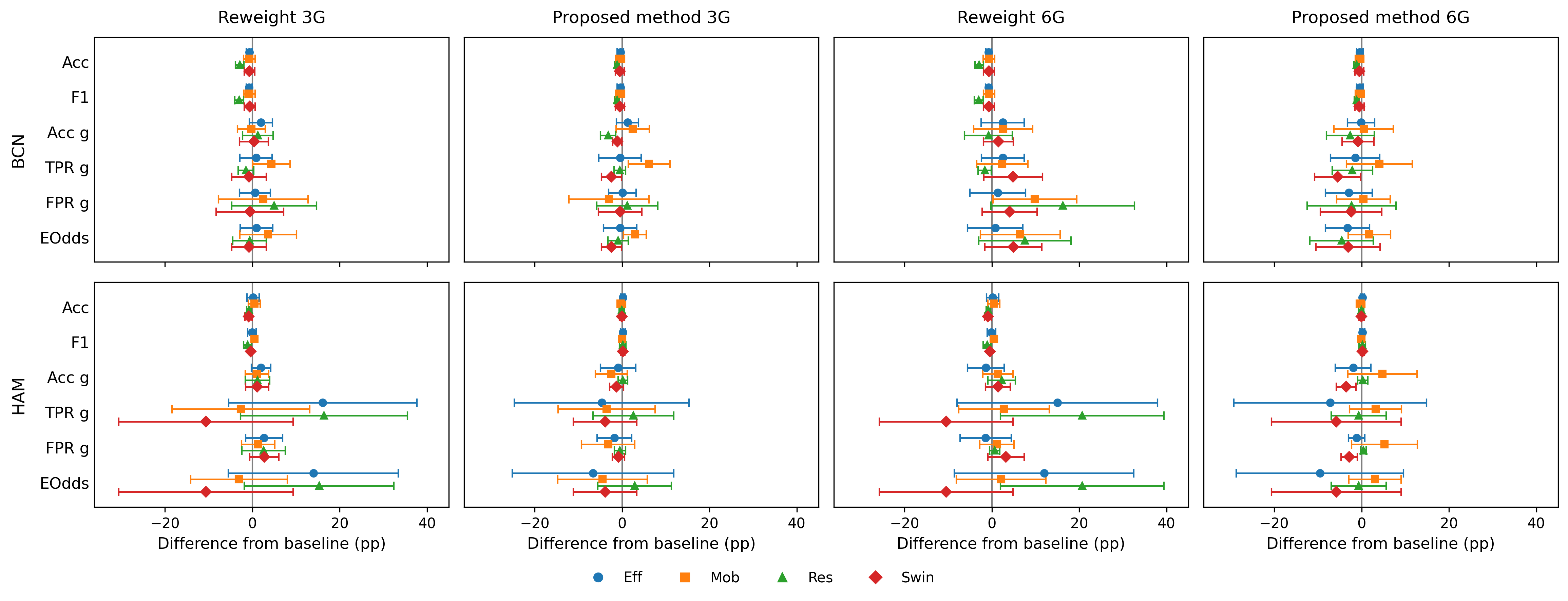}
    \caption{Changes in performance and fairness metrics from the baseline for reweight and our proposed method. The points represent the average of 5 seeds, and the error bars indicate the 95\% CI.}
    \label{fig:overall_results}
\end{figure*}

Figure \ref{fig:overall_results} shows the changes in performance and fairness compared to the baseline for the proposed method (WD) and Reweight. The vertical gray lines indicate a difference of 0 pp from the baseline; for Accuracy and F1, a positive direction indicates an improvement in performance, while for the fairness gap, a negative direction indicates a reduction in the disparity. Regarding classification performance, WD maintained performance close to the baseline for most models in both the 3-group and 6-group evaluations, showing a tendency toward smaller performance degradation compared to Reweight. For fairness metrics, WD generally concentrated around 0 or shifted in the negative direction, demonstrating more stable reduction of disparities than the categorical reweighting. In the 6-group evaluation, with Reweight, the TPR and Equalized-Odds gap shifted significantly in the positive direction in ResNet50 and EfficientNet-B0, revealing unstable behavior with fine-grained categorization. In contrast, many models of WD (except for MobileNetV2) tended to narrow fairness gaps while maintaining performance to a large extent. Although improvements were inconsistent in MobileNetV2, EfficientNet-B0—another lightweight CNN—exhibited the same improvement trend as the other models. This suggests that this behavior cannot be explained by model size and points to an interaction with MobileNetV2’s specific architecture. These results imply that when continuous sensitive attributes, such as skin tone, are discretized into categories, the effectiveness of bias correction can become unstable depending on how category boundaries are set. WD, which adjusts samples individually in a continuous manner, has been shown to achieve a more stable fairness–utility trade-off. \\

\noindent\textbf{Limitations:} HAM synthetized color and luminance \cite{tschandl2018ham10000}. It may affect the ITA distribution, but our analysis relies on relative distributional differences rather than absolute ITA, limiting its impact. The underrepresentation of darker skin color may lead to unstable performance estimates and limit the generalizability of our findings. The results of BCN can be more reliable. Future work can validate this on more diverse datasets.

\section{Conclusion}
This study focused on skin tone bias in skin lesion classification and proposed a fairness evaluation and mitigation method that treats skin tone as a continuous sensitive attribute. Experimental results confirmed performance disparities and the proposed method demonstrated more stable improvements in fairness while generally maintaining diagnosis performance. These results suggest the method’s utility in realizing medical image AI that is more robust and fair with respect to skin tone diversity.

\section{Compliance with Ethical Standards}
This study used publicly available HAM10000 and BCN20000 datasets, whose original collection received ethical approval. No new human participants were involved.
\section{Acknowledgments}
No funding was received. The authors declare no conflicts of interest. We thank the EPSRC National Edge AI Hub for its support. DeepL and Grammarly assisted with translation and proofreading.

\bibliographystyle{IEEEbib}
\bibliography{main}

\end{document}